\documentclass[10pt,twocolumn,letterpaper]{article}

\usepackage{wacv}
\usepackage{times}
\usepackage{epsfig}
\usepackage{graphicx}
\usepackage{amsmath}
\usepackage{amssymb}
\usepackage{comment}

\usepackage[pagebackref=true,breaklinks=true,letterpaper=true,colorlinks,bookmarks=false]{hyperref}

\wacvfinalcopy

\def\wacvPaperID{470}

\ifwacvfinal\fi
\def\eg{\emph{e.g.}}

\def\etal{\emph{et al.}}
\def\ie{\emph{i.e.}}

  \usepackage{tikz,pgfplots}
  \usetikzlibrary{plotmarks,shapes,arrows}
  \pgfplotsset{compat=newest}

  \usepackage{amsmath,amssymb,amsfonts,amsthm,mathrsfs}

  \newcommand{\T}{^\mathsf{T}}
  
  \newcommand{\R}{\mathbb{R}}

  \providecommand{\norm}[1]{\|#1\|}

  \newcommand{\N}{\mathrm{N}}

  \usepackage{bm}
  
  \newcommand{\mbf}[1]{\mathbf{#1}}
  \newcommand{\vect}[1]{\mbf{#1}}
  \newcommand{\vectb}[1]{\bm{#1}}

  \usepackage{xcolor}

  \usepackage{graphicx}
  \graphicspath{ {./}{./fig/} }

  \usepackage{subcaption}

  \usepackage[noend,ruled,noline]{algorithm2e}

  \usepackage{booktabs}

  \newlength\figureheight
  \newlength\figurewidth

  \usepackage{color}

  \usepackage[square,numbers,sort&compress]{natbib}

  \DeclareRobustCommand{\marker}[1]{\tikz[baseline=(char.base)]{%
    \node[shape=circle,draw,inner sep=1pt, thin, black!70, outer sep=0, black, fill=white] (char) {\sffamily\small#1};}}

\makeatletter
\newenvironment{customlegend}[1][]{%
    \begingroup

    \pgfplots@init@cleared@structures
    \pgfplotsset{#1}%
}{%

    \pgfplots@createlegend
    \endgroup
}%

\def\addlegendimage{\pgfplots@addlegendimage}
\makeatother

\ifwacvfinal\fi
\begin{document}

\title{PIVO: Probabilistic Inertial-Visual Odometry for Occlusion-Robust Navigation}

\author{Arno Solin \\
Aalto University\\
{\tt\small\scalebox{.92}[1.0]{arno.solin@aalto.fi}}
\and
Santiago Cort\'es \\
Aalto University\\
{\tt\small\scalebox{.92}[1.0]{santiago.cortesreina}} \\
{\tt\small\scalebox{.92}[1.0]{@aalto.fi}} \vspace*{-14pt}
\and
Esa Rahtu \\
Tampere Univ.\ of Tech.\\
{\tt\small\scalebox{.92}[1.0]{esa.rahtu@tut.fi}}
\and
Juho Kannala \\
Aalto University\\
{\tt\small\scalebox{.92}[1.0]{juho.kannala@aalto.fi}}
}

\maketitle
\ifwacvfinal\thispagestyle{empty}\fi

\begin{abstract}
  This paper presents a novel method for visual-inertial odometry. The method is based on an information fusion framework employing low-cost IMU sensors and the monocular camera in a standard smartphone. We formulate a sequential inference scheme, where the IMU drives the dynamical model and the camera frames are used in coupling trailing sequences of augmented poses. The novelty in the model is in taking into account all the cross-terms in the updates, thus propagating the inter-connected uncertainties throughout the model. Stronger coupling between the inertial and visual data sources leads to robustness against occlusion and feature-poor environments. We demonstrate results on data collected with an iPhone and provide comparisons against the Tango device and using the EuRoC data set.
\end{abstract}

\section{Introduction}
Accurate tracking of the orientation and location of mobile devices is important for many applications. Examples include robot navigation, pedestrian navigation and wayfinding, augmented reality, games, safety and rescue (\eg\ for firefighters). Motivated by the practical relevance of the problem, there is plenty of recent research on device tracking based on various kinds of sensors. Perhaps one of the most promising approaches for precise real-time tracking is visual-inertial odometry, which is based on fusing measurements from inertial sensors (\ie\ an accelerometer and gyroscope) with visual feature tracking from video. This kind of approach is practical since both video cameras and MEMS-based inertial measurement units are nowadays low-cost commodity hardware which are embedded in most smartphones and tablet devices. Further, the two sensor modalities complement each other; visual odometry can provide good precision in visually distinguishable environments whereas inertial navigation is robust to occlusion and able to provide absolute metric scale for the motion.

Visual-inertial odometry methods have already been demonstrated in many research papers and in some commercial devices, such as the Tango device by Google. However, most published methods are heavily camera-based in the sense that they need unobstructed visibility and plenty of visual features in the scene all the time. For example, the tracking of the Tango device fails if the camera lens is fully occluded. Further, since pure inertial navigation has been considered challenging or impossible with low-cost consumer grade inertial sensors \cite{Sachs:2010}, so far there are no occlusion-robust visual-inertial odometry demonstrated using conventional smartphone hardware.

\begin{figure*}

  \pgfplotsset{
    trim axis right,
    yticklabel style={rotate=90},
  }

  \setlength{\figurewidth}{1.09\textwidth}

  \footnotesize\centering
  \input{./fig/city-wide-navigation-rot90.tex}

  \includegraphics[width=.123\textwidth]{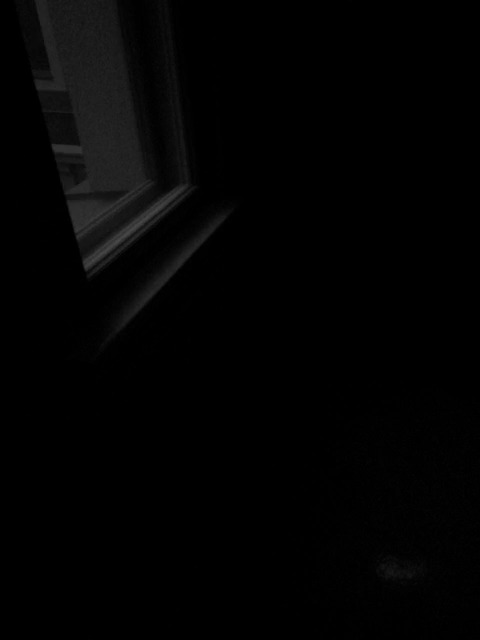}\hfill%
  \includegraphics[width=.123\textwidth]{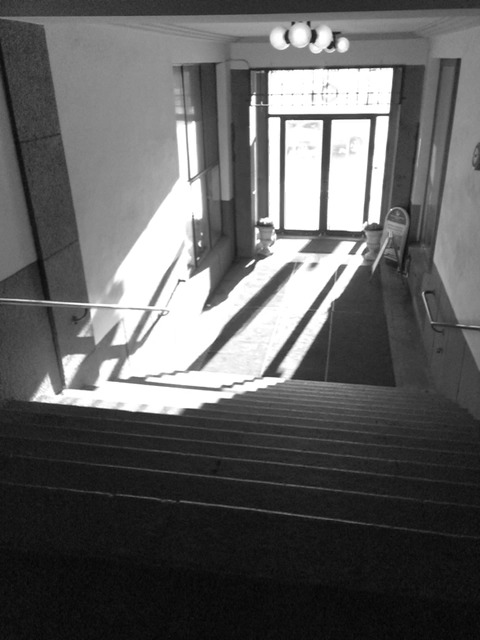}\hfill%
  \includegraphics[width=.123\textwidth]{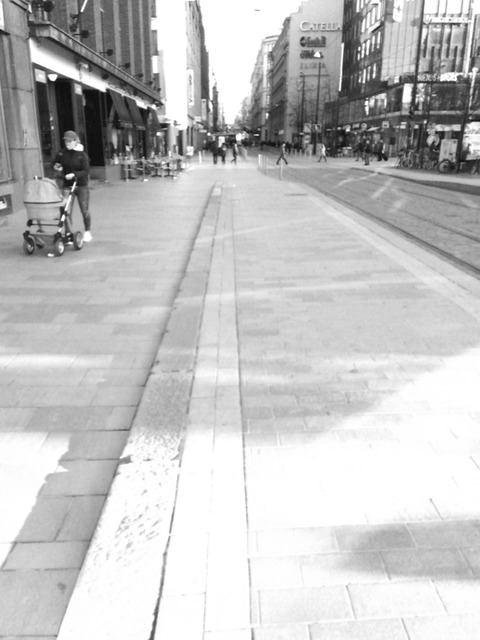}\hfill%
  \includegraphics[width=.123\textwidth]{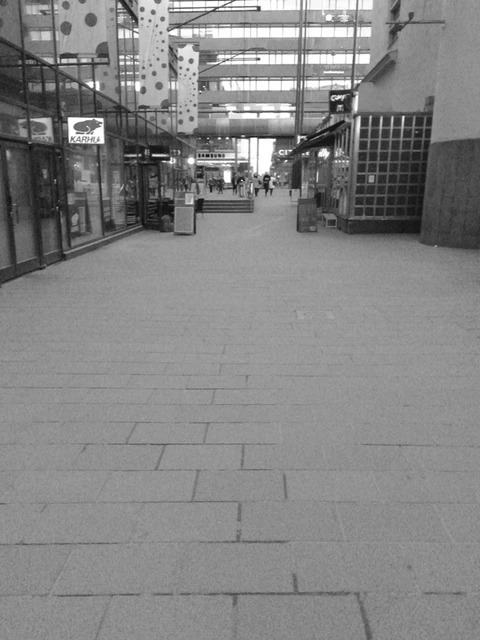}\hfill%
  \includegraphics[width=.123\textwidth]{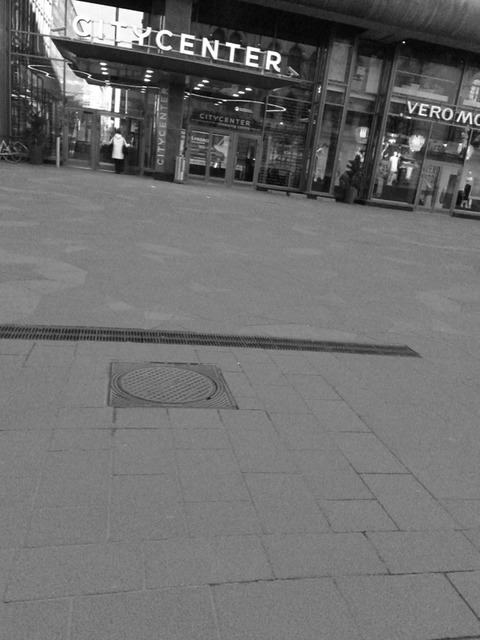}\hfill%
  \includegraphics[width=.123\textwidth]{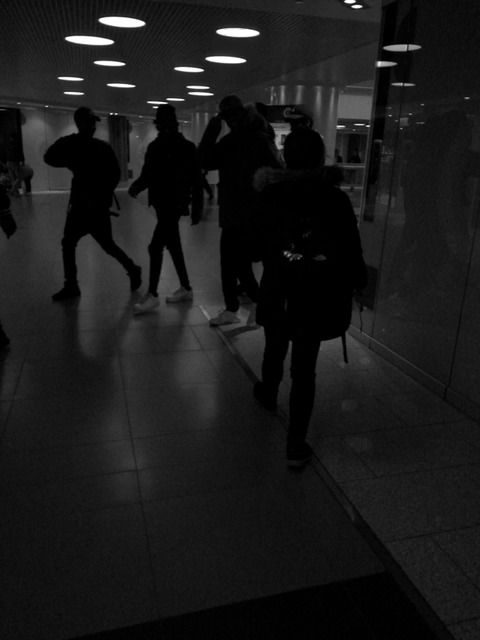}\hfill%
  \includegraphics[width=.123\textwidth]{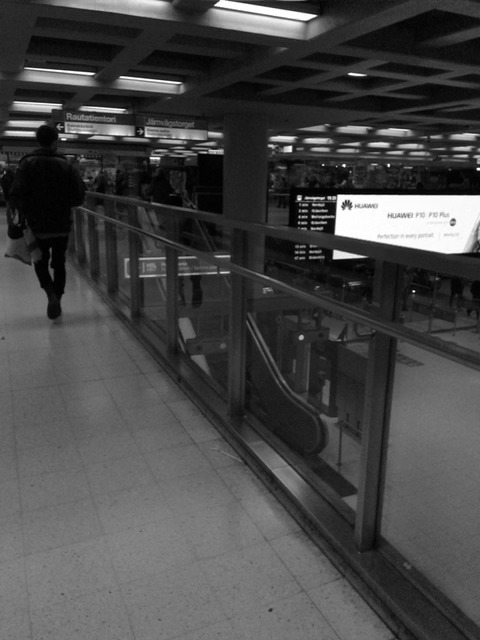}\hfill%
  \includegraphics[width=.123\textwidth]{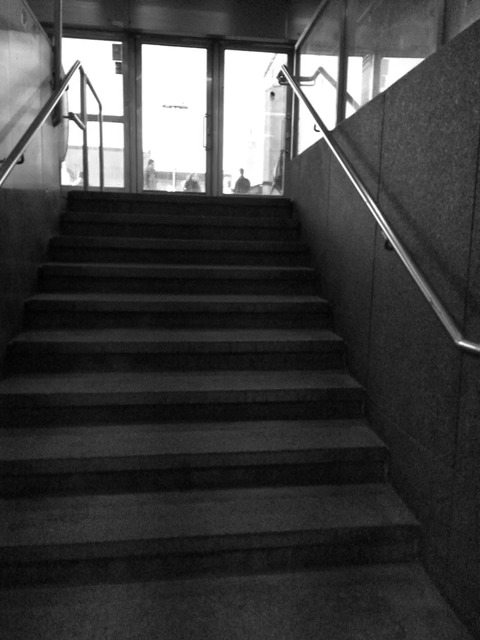}\hfill%
  \\
  \caption{PIVO tracking on a smartphone (iPhone~6) starting from an office building (1--2), through city streets (3--5), a shopping mall (6), and underground transportation hub (7--8).}
  \label{fig:example}
  \vspace{-1em}
\end{figure*}

In this paper, we propose a probabilistic inertial-visual odometry technique which allows occlusion-robust navigation on standard smartphone hardware. Our approach is able to utilize consumer-grade MEMS-based inertial sensors in a more robust manner than previous approaches and we therefore call our approach \emph{inertial-visual} odometry instead of the commonly used term `visual-inertial' odometry. By taking the full advantage of the inertial sensors we are able to demonstrate precise odometry with a standard iPhone in use cases where the camera is temporarily fully occluded (\eg\ in a bag or by people in a crowd) for extended periods of time. A visual example of such a path is shown in Figure~\ref{fig:example}.

We aim at maximizing the information throughput from both the visual and inertial modality. This means both accurate tracking and insensitivity to partial and complete occlusion for extended periods of time. More exactly, this goal is covered by the  following three contributions of this paper novel to visual-inertial schemes: 
  (i)~An efficient IMU propagation model in discrete-time, which keeps the error sources limited to the linearization error; 
  (ii)~A trailing pose augmentation through a series of linear Kalman updates which preserve the cross-covariance between current and past poses; 
  (iii)~A visual update step where the estimated feature point coordinate is integrated out, and thus the only approximation error comes from the linearization in the EKF.
For all of the methodology a core design criterion is formulating the methods in sequential manner suitable for real-time running on current mobile hardware and building upon noisy sensor sources available in standard smartphones.

\section{Related work}
\label{sec:background}
Methods for tracking the pose of a mobile device with six degrees of freedom can be categorized based on {(i)}~the input sensor data, or {(ii)}~the type of the approach. Regarding the latter aspect the main categories are simultaneous localization and mapping (SLAM) approaches, which aim to build a global map of the environment and utilize it for loop-closures and relocalization, and pure odometry approaches, which aim at precise sequential tracking without building a map or storing it in memory. SLAM is particularly beneficial in cases where the device moves in a relatively small environment and revisits the mapped areas multiple times, since the map can be used for removing the inevitable drift of tracking. However, accurate low-drift odometry is needed in cases where the device moves long distances without revisiting mapped areas.

Regarding the types of input data there is plenty of literature using various combinations of sensors. Monocular visual SLAM and odometry techniques, which use a single video camera, are widely studied \cite{Engel+Schops+Cremers:2014, Forster+Pizzoli+Scaramuzza:2014,Klein+Murray:2009} but they have certain inherent limitations which hamper their practical use. That is, one can not recover the absolute metric scale of the scene with a monocular camera and the tracking breaks if the camera is occluded or there are not enough visual features visible all the time. For example, homogenous textureless surfaces are quite common in indoor environments but lack visual features. Moreover, even if the metric scale of the device trajectory would not be necessary in all applications, monocular visual odometry can not keep a consistent scale if the camera is not constantly translating---that is, pure rotations cause problems \cite{Herrera+Kim+Kannala+Pulli+Heikkila:2014}. Scale drift may cause artifacts even without pure rotational motion if loop-closures can not be frequently utilized \cite{Strasdat+Montiel+Davison:2010}. 

Methods that use stereo cameras are able to recover the metric scale of the motion, and consistent tracking is possible even if the camera rotates without translating \cite{Mur-Artal+Tardos:2016,Engel+Stuckler+Cremers:2015}. Still, the lack of visually distinguishable texture and temporary occlusions (\eg\ in a crowd) are problem for all approaches that utilize only cameras. Recently, due to the increasing popularity of depth sensing RGB-D devices (either utilizing structured light or time-of-flight), also SLAM and odometry approaches have emerged for them \cite{Izadi+Newcombe+Kim+Hilliges+Molyneaux+Hodges+Kohli+Shotton+Davison+Fitzgibbon:2011, Newcombe+Fox+Seitz:2015, Kerl+Stuckler+Cremers:2015}. These devices provide robustness to lack of texture but they also require unobstructed line of sight, and hence occlusions may still be a problem. In addition, many of the cameras have a limited range for depth sensing and do not work outdoors because they utilize infrared projectors.

Thus, in order to make tracking more robust and practical for consumer applications on mobile devices both indoors and outdoors, it has become common to combine video cameras with inertial measurement units (IMUs) \cite{Mourikis+Roumeliotis:2007,Hesch+Kottas+Bowman+Roumeliotis:2014, Blosch+Omari+Hutter+Siegwart:2015, Tanskanen+Naegeli+Pollefeys+Hilliges:2015, Mur-Artal+Tardos:2017b, Forster+Carlone+Dellaert+Scaramuzza:2017}. Examples of hardware platforms that provide built-in visual-inertial odometry are Google Tango and Microsoft Hololens devices. However, both of these devices contain custom hardware components (\eg\ a fisheye lens camera), which are not common in conventional smartphones. In addition, there are several research papers which utilize IMUs with custom stereo camera setups \cite{Usenko+Engel+Stuckler+Cremers:2016, Leutenegger+Lynen+Bosse+Siegwart+Furgale:2015, Forster+Zhang+Gassner+Werlberger+Scaramuzza:2017,Heng+Choi:2016}. 

In this paper we present a probabilistic approach for fusing information from consumer grade inertial sensors (\ie\ 3-axis accelerometer and gyroscope) and a monocular video camera for accurate low-drift odometry. This is practically the most interesting hardware setup as most modern smartphones contain a monocular video camera and an IMU. Despite the wide application potential of such hardware platform, there are not many previous works which demonstrate visual-inertial odometry using standard smartphone sensors. This is most likely due to the relatively low quality of low-cost IMUs which makes inertial navigation challenging. The most notable papers covering the smartphone use case are \cite{Klein+Murray:2009,Li+Kim+Mourikis:2013, Tanskanen+Kolev+Meier+Camposeco+Saurer+Pollefeys:2013}. However, all these previous approaches are either \emph{visual-only} or \emph{visual-heavy} in the sense that tracking breaks if there is complete occlusion of camera for short periods of time. This is the case also with the visual-inertial odometry of the Google Tango device.

\section{Inertial-visual information fusion}
\label{sec:materials-and-methods}
Consider a device with a monocular camera, an IMU with 3-axis gyroscope/accelerometer, and known camera-to-IMU translational and rotational offsets---a characterization that matches modern day smartphones. In the following, we formulate the PIVO approach for fusing information from these data sources such that we maintain all dependencies between uncertain information sources---up to the linearization error from the non-linear filtering approach. The outline of the PIVO method is summarized in Algorithm~\ref{alg:pseudocode}.

\subsection{Non-linear filtering for information fusion}
In the following, we set the notation for the non-linear filtering approach (see \cite{Sarkka:2013} for an overview) for information fusion in the paper. We are concerned with non-linear state-space equation models of form
\begin{align}
  \vect{x}_k &= \vect{f}_k(\vect{x}_{k-1}, \vectb{\varepsilon}_k), \label{eq:dynamic} \\
  \vect{y}_k &= \vect{h}_k(\vect{x}_k) + \vectb{\gamma}_k, \label{eq:measurement} 
\end{align}
where $\vect{x}_k \in \R^n$ is the state at time step $t_k$, $k=1,2,\ldots$, $\vect{y}_k \in \R^m$ is a measurement, $\vectb{\varepsilon}_k \sim \N(\vectb{0}, \vect{Q}_k)$ is the Gaussian process noise, and  $\vectb{\gamma}_k \sim \N(\vectb{0},\vect{R}_k)$ is the Gaussian measurement noise. The dynamics and measurements are specified in terms of the dynamical model function $\vect{f}_k(\cdot, \cdot)$ and the measurement model function $\vect{h}_k(\cdot)$, both of which can depend on the time step $k$.
The extended Kalman filter (EKF, \cite{Jazwinski:1970, Bar-Shalom+Li+Kirubarajan:2001, Sarkka:2013}) provides a means of approximating the state distributions $p(\vect{x}_k \mid \vect{y}_{1:k})$ with Gaussians:
  $p(\vect{x}_k \mid \vect{y}_{1:k}) \simeq \N(\vect{x}_k \mid \vect{m}_{k \mid k}, \vect{P}_{k \mid k})$.

The linearizations inside the extended Kalman filter cause some errors in the estimation. Most notably the estimation scheme does not preserve the norm of the orientation quaternions. Therefore after each update an extra quaternion normalization step is added to the estimation scheme.

In case either the dynamical \eqref{eq:dynamic} or measurement model~\eqref{eq:measurement} is linear (\ie\ $\vect{f}_k(\vect{x},\vectb{\varepsilon}) = \vect{A}_k \, \vect{x} + \vectb{\varepsilon}$ or $\vect{h}_k(\vect{x}) = \vect{H}_k \, \vect{x}$, respectively), the prediction/update steps reduce to the closed-form solutions given by the conventional Kalman filter.

\begin{algorithm}[t!]
  \caption{Outline of the PIVO method.}

  \label{alg:pseudocode}
  \footnotesize
  \DontPrintSemicolon
  \SetCommentSty{it}
  \SetKwComment{tcp}{}{}
  Initialize the state mean and covariance\;
  \ForEach{IMU sample pair $(\vect{a}_k,\vectb{\omega}_k)$}{
    Propagate the model with the IMU sample \tcp*{see Sec.~\ref{sec:IMU-propagation}}
    Perform the EKF prediction step\;
    \If{new frame is available}{
      track visual features\;
      \ForEach{feature track}{
        Jointly triangulate feature using poses in state and 
          calculate the visual update proposal \tcp*{see Sec.~\ref{sec:visual-update}}
        \If{proposal passes check}{
          Perform the EKF visual update\;
        }
      }
      Update the trail of augmented poses \tcp*{see Sec.~\ref{sec:pose-augmentation}}
    }
  }

\end{algorithm}

\subsection{IMU propagation model}
\label{sec:IMU-propagation}
The state variables of the system hold the information of the current system state and a fixed-length window of past poses in the IMU coordinate frame:
\begin{equation}
  \vect{x}_k = (\vect{p}_k, \vect{q}_k, \vect{v}_k, \vect{b}_k^\mathrm{a}, \vect{b}_k^{\omega}, \vect{T}_k^\mathrm{a}, \vectb{\pi}^{(1)}, \vectb{\pi}^{(2)}, \ldots, \vectb{\pi}^{(n_\mathrm{a})}),
\end{equation}
where $\vect{p}_k \in \R^3$ is the device position, $\vect{v}_k \in \R^3$ the velocity, and $\vect{q}_k \in \R^4$ the orientation quaternion at time step $t_k$. Additive accelerometer and gyroscope biases are denoted by $\vect{b}_k^\mathrm{a} \in \R^3$ and $\vect{b}_k^{\omega} \in \R^3$, respectively. $\vect{T}_k^\mathrm{a} \in \R^{3 \times 3}$ holds the multiplicative accelerometer bias. The past device poses are kept track of by augmenting a fixed-length trail of poses,  $\{\vectb{\pi}^{(i)}\}_{i=1}^{n_\mathrm{a}}$, where $\vectb{\pi}^{(i)} = (\vect{p}_i, \vect{q}_i)$, in the state. 

Contrary to many previous visual-inertial methods, we seek to define the propagation method directly in discrete-time following Solin~\etal\ \cite{Solin+Cortes+Rahtu+Kannala}. The benefits are that the derivatives required for the EKF prediction are available in closed-form, no separate ODE solver iteration is required, and possible pitfalls (see, \eg, \cite{Sarkka+Solin:2012}) related to the traditional continuous-discrete formulation \cite{Jazwinski:1970} can be avoided.

The IMU propagation model is given by the mechanization equations
\begin{equation}
  \begin{pmatrix}
    \vect{p}_k \\ \vect{v}_k \\ \vect{q}_k
  \end{pmatrix}
  =
  \begin{pmatrix}
    \vect{p}_{k-1} + \vect{v}_{k-1}\Delta t_k \\
    \vect{v}_{k-1} + [\vect{q}_k (\tilde{\vect{a}}_k + \vectb{\varepsilon}^\mathrm{a}_k) \vect{q}_k^\star - \vect{g}] \Delta t_k \\
    \vectb{\Omega}[(\tilde{\vectb{\omega}}_k + \vectb{\varepsilon}^\omega_k) \Delta t_k] \vect{q}_{k-1}
  \end{pmatrix},
\end{equation}
where the time step length is given by $\Delta t_k = t_{k} - t_{k-1}$, the acceleration input is $\tilde{\vect{a}}_k \in \R^3$ and the gyroscope input by $\tilde{\vectb{\omega}}_k \in \R^3$. Gravity $\vect{g}$ is a constant. The quaternion rotation is denoted by $\vect{q}_k [\cdot] \vect{q}_k^\star$, and the quaternion update is given by $\vectb{\Omega}: \R^3 \to \R^{4 \times 4}$ (see, \eg, \cite{Titterton+Weston:2004}). 
The process noises associated with the IMU data are treated as i.i.d.\ Gaussian noise $\vectb{\varepsilon}^\mathrm{a}_k \sim \N(\vectb{0},\vectb{\Sigma}^\mathrm{a} \Delta t_k)$ and $\vectb{\varepsilon}^\omega_k \sim \N(\vectb{0},\vectb{\Sigma}^\omega \Delta t_k)$.

Low-cost IMU sensors suffer from misalignment and scale errors in addition to white measurement noise. These are accounted for in the dynamic model by including estimation for multiplicative and additive scale bias for the accelerometer, $\tilde{\vect{a}}_k = \vect{T}_k^\mathrm{a} \, \vect{a}_k - \vect{b}^\mathrm{a}_k$, and an additive bias for the gyroscope, $\tilde{\vectb{\omega}}_k = \vectb{\omega}_k - \vect{b}^\omega_k$, where $\vect{a}_k$ and $\vectb{\omega}_k$ are the accelerometer and gyroscope sensor readings at $t_k$, and $\vect{b}^\mathrm{a}_k$ and $\vect{b}^\omega_k$ are the additive biases. Miscalibrations in the accelerometer scale are handled by the diagonal scale error matrix $\vect{T}_k^\mathrm{a}$.

In this work, we ignore the slow random walk of the sensor biases, and thus the biases and scale error terms are considered fixed and estimated as a part of the state. Similarly, the augmented poses $\vectb{\pi}^{(i)}$ also have no temporal dynamics.
The complete dynamical model must be differentiated both in terms of the state variables and process noise terms for the EKF estimation scheme. The derivatives are available in closed-form, which helps preserving the numerical stability of the system.
As a part of the estimation setup we deal with zero-velocity and velocity pseudo-measurement updates similarly as in \cite{Solin+Cortes+Rahtu+Kannala}.

\subsection{Camera pose augmentation}
\label{sec:pose-augmentation}
Once a new camera frame is acquired at time $t_\star$ (the $t_\star$ is matched against the IMU clock such that $t_\star$ is assigned the most recent $t_k$), the current camera pose given by the state variables $\vectb{\pi}^\star = (\vect{p}_\star, \vect{q}_\star)$ is augmented into the state by a linear Kalman update. 

The camera pose augmentation is a two-step process. First, the pose trail propagation step takes care of moving the past poses forward, discarding the oldest pose, and initiating a prior for the current pose. As the state distributions are all Gaussians, in probabilistic sense this corresponds to the following linear Kalman prediction step with the following dynamic model
\begin{equation} \label{eq:augmentation_dyn}
  \vect{A}_\star = 
  \begin{pmatrix}
    \vect{I}_{19} & \vectb{0} & \vectb{0} \\
    \vectb{0} & \vectb{0} & \vectb{0} \\
    \vectb{0} & \vect{I}_{7(n_\mathrm{a}-1)} & \vectb{0}
  \end{pmatrix}
\end{equation}
and $\vect{Q}_\star = \mathrm{blkdiag}(\vectb{0}_{19}, \sigma_\mathrm{p}^2\,\vect{I}_3, \sigma_\mathrm{q}^2\,\vect{I}_4, \vectb{0}_{7(n_\mathrm{a}-1)})$, where the prior variances for the new pose position and orientation, $\sigma_\mathrm{p}^2$ and $\sigma_\mathrm{q}^2$, are set large to keep the prior uninformative. $\vect{I}_d$ and $\vectb{0}_d$ denote identity and zero matrices with dimension $d$, respectively.

The second step of the pose augmentation takes care of including the current system pose frontmost in the pose history trail in the state. As the state distribution is Gaussian, the state update corresponds to a linear Kalman update step with the following measurement model matrix
\begin{equation} \label{eq:augmentation_meas}
  \vect{H}_\star = 
  \begin{pmatrix}
    \vect{I}_{7} & \vectb{0}_{7 \times 19} & -\vect{I}_{7} & & \vectb{0}_{7 \times 7(n_\mathrm{a}-1)}
  \end{pmatrix}
\end{equation}
and a measurement noise covariance $\vect{R} = \sigma^2_\star\,\vect{I}$ very close to zero and representing the small residual error related to the misalignment of the sensor and frame timestamps. Setting $\vect{y}_\star = \vectb{0}$, defines a Kalman update that enforces the current pose to match the first augmented state slot $\vectb{\pi}^{(1)}$.

The length of the augmented pose trail $n_\mathrm{a}$ is a selectable parameter, it can be tuned according to the available computing resources or the accuracy requirements. In theory the length of the filter state as well as the frame rate of the camera could also be adaptively controlled during the estimation.

Even though the state only holds a fixed number of poses and the oldest pose is discarded, formally the poses are not removed from the model. Our formulation means that the result is the same as it would have been if \emph{every pose} would remain in the state, even though the information is used only in the trailing window. This would of course be computationally infeasible and totally unnecessary as the information is not used later on. However, it means that the state variables are altered only in a fashion that preserves the cross-covariances between future and past states.

\subsection{Visual update model}
\label{sec:visual-update}
\begin{figure}[!t]

  \pgfplotsset{
    trim axis right,
    yticklabel style={rotate=90},
  }
  \begin{subfigure}[b]{.49\textwidth}
    \footnotesize\centering%

    \setlength{\figurewidth}{1.2\textwidth}
    \setlength{\figureheight}{0.80\figurewidth}
    \hspace*{2em}%
    \input{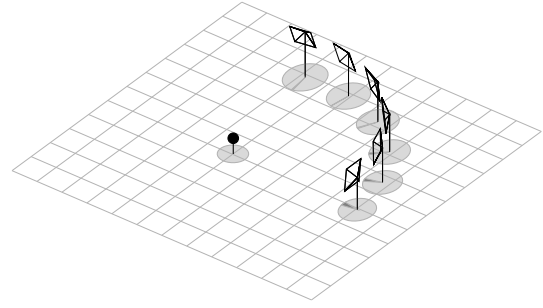}

    \caption{Poses and feature estimates}
    \label{fig:update-sketch}
  \end{subfigure}
  \\[1em]
  \begin{subfigure}[b]{.49\textwidth}
    \footnotesize\centering%

    \setlength{\figureheight}{.67\textwidth}%
    \setlength{\figurewidth}{1.3345\figureheight}
    \hspace*{-2em}%
    {\tiny\input{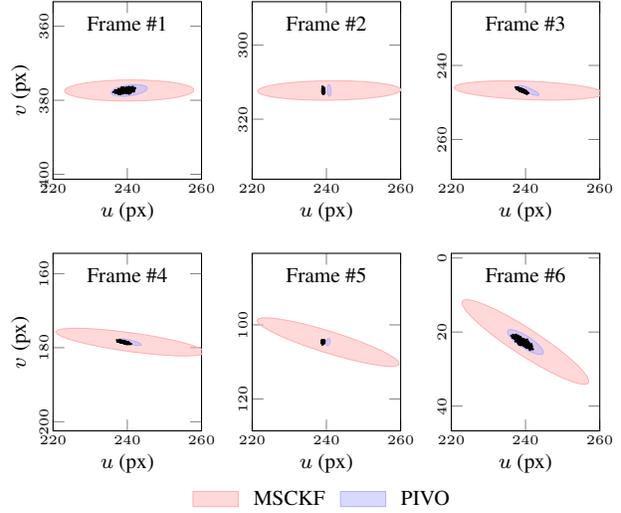}}
    \hspace*{2.3cm}%
    \begin{tikzpicture}
      \begin{customlegend}[legend columns=-1,legend style={draw=none,column sep=1ex},legend entries={MSCKF, PIVO}]
        \addlegendimage{red!30,fill=red!15,area legend}
        \addlegendimage{blue!30,fill=blue!15,area legend}
      \end{customlegend}
    \end{tikzpicture}
    \caption{Comparison between PIVO and MSCKF visual update model}
    \label{fig:comparison-MSCKF-PIVO}
  \end{subfigure}
  \caption{(a)~A visual feature observed by a trail of camera poses with associated uncertainties. In (b) the black dots are the `true' distributions for the visual update model. The red patch shows shape of the Gaussian approximation used by the MSCKF visual update, and blue shows the shape of the approximation used by PIVO.}
  \label{fig:visual}
  \vspace{-1em}
\end{figure}

The visual measurement model formulation bears resemblance to the model used in MSCKF \cite{Mourikis+Roumeliotis:2007}, but the linearization for the EKF update is handled differently. The motivation is that tracking a visual feature over several camera frames imposes a long time-range constraint on the device movement together with the camera intrinsics. What we underline is that not only the feature points, but also all inter-connected uncertainties between \emph{all} state variables need to be bundled in the update. See Figure~\ref{fig:update-sketch} for an outline of the elements involved in the visual update model. This makes the derivation of the visual update step slightly more complicated than that in \cite{Mourikis+Roumeliotis:2007}.

For feature tracking, we use the Shi--Tomasi \emph{Good features to track} approach \cite{Shi+Tomasi:1994} to determine strong corners in the image. These features are tracked across frames by a pyramidal Lucas--Kanade tracker \cite{Lucas+Kanade:1981,Bouguet:2001}. Lost features are replaced by new features and with preference of having the features evenly scattered over the field of view.

In this work the visual update is performed per tracked feature, which keeps the computational complexity constant per feature. The feature update for a given track $j$ of feature point pixel coordinates $\vect{y}^{(j)} = (u_1, v_1, u_2, v_2, \ldots, u_m, v_m)$ seen in $m$ past frames with augmented poses $\vectb{\pi}^{(i)} = (\vect{p}^{(i)}, \vect{q}^{(i)})$ still in the state ($m \leq n_\mathrm{a}$).

We construct a `full' measurement model representing any projection of an arbitrary feature point coordinate to the observed image pixel coordinates:
\begin{equation}\label{eq:visual-update}
  \vect{y}_{k}^{(j)} = \vect{h}_k^{(j)}(\vect{x}) + \vectb{\gamma}_k, \quad \vectb{\gamma}_k \sim \N(\vectb{0},\sigma^2_\mathrm{uv}\,\vect{I}_{2m}),
\end{equation}
where $\sigma_\mathrm{uv}$ is the measurement noise standard deviation in pixel units. In the following we drop the time index subscript $k$ and will write out the function $\vect{h}$ per frame $i$ (\ie\ for pixel pairs $\vect{y}_i = (u_i,v_i)$ in $\vect{y}$). In our formulation the feature global coordinate $\vect{p}_*^{(j)} \in \R^3$ will be integrated out in the final model, which differs from previous approaches. We, however, write out the derivation of the model by including the estimation of $\vect{p}_*^{(j)}$.
\begin{equation}\label{eq:visual-update-parts}
  \vect{h}^{(j, i)}(\vect{x}) = \vect{g}\big(\vect{R}({^\mathrm{C}}\vect{q}^{(i)}) \, (\vect{p}_*^{(j)} - {^\mathrm{C}}\vect{p}^{(i)})\big),
\end{equation}
where the rotation and translation in the global frame corresponds to the camera extrinsics calculated from the device pose and known rotational and translational offsets between IMU and camera coordinate frames (denoted by the superscript `C' in Eq.~\ref{eq:visual-update-parts}). The camera projection is modeled by a standard perspective model $\vect{g}:\R^3 \to \R^2$ with radial and tangential distortion \cite{Heikkila+Silven:1997} and calibrated off-line.

To estimate the position $\vect{p}_*^{(j)}$ of a tracked feature we employ a similar approach as in \cite{Mourikis+Roumeliotis:2007}, where the following minimization problem is set up:
  $\vectb{\theta}_* = \mathrm{arg\,min}_{\vectb{\theta}} \sum_{i=1}^m \norm{\vectb{\varphi}_i(\vectb{\theta})}$,
where we use the inverse depth parametrization, $\vectb{\theta} = 1/p_{\mathrm{z}}(p_{\mathrm{x}}, p_{\mathrm{y}}, 1)$, to avoid local minima and improve numerical stability \cite{Montiel+Civera+Davison:2006}. The target function $\vectb{\varphi}_i : \R^3 \to \R^2$ can be defined as follows on a per frame basis:
\begin{align}
  \vectb{\varphi}_i(\vectb{\theta}) &= \vect{\tilde y}_i - h_{i,3}^{-1} \begin{pmatrix} h_{i,1} & h_{i,2} \end{pmatrix}\T, \\
  \vect{h}_i &= \vect{C}_i \begin{pmatrix} \theta_1 & \theta_2 & 1 \end{pmatrix}\T + \theta_3 \, \vect{t}_i, \\
  \vect{C}_i &= \vect{R}({^\mathrm{C}}\vect{q}^{(i)}) \, \vect{R}\T({^\mathrm{C}}\vect{q}^{(1)}), \\
  \vect{t}_i &= \vect{R}({^\mathrm{C}}\vect{q}^{(i)}) \, \big(\vect{p}^{(1)} - \vect{p}^{(i)}\big),
\end{align}
where feature pixel coordinates $\vect{\tilde y}_i$ are undistorted from $\vect{y}_i$. For solving the minimization problem a Gauss--Newton minimization scheme is employed:
\begin{equation}
  \vectb{\theta}^{(s+1)} = \vectb{\theta}^{(s)} - ( \vect{J}_{\vectb{\varphi}}\T \, \vect{J}_{\vectb{\varphi}} )\T \, \vect{J}_{\vectb{\varphi}}\T \, \vectb{\varphi}(\vectb{\theta}^{(s)}),
\end{equation}
where $\vect{J}_{\vectb{\varphi}}$ is the Jacobian of $\vectb{\varphi}$. The iteration is initialized by an intersection estimate $\vectb{\theta}^{(0)}$ calculated just from the first and last pose.

The beef of this section is that in order to do a precise EKF update with measurement model \eqref{eq:visual-update} the entire procedure described after Equation~\eqref{eq:visual-update} needs to be differentiated in order to derive the closed-form Jacobian $\vect{H}_{\vect{x}}:\R^n \to \R^{2m \times n}$. This includes differentiating the entire Gauss--Newton scheme iterations with respect to \emph{all} state variables.

The effect of taking into account all the cross-derivatives is illustrated in Figure~\ref{fig:comparison-MSCKF-PIVO}. The figure illustrates how the visual update for a three frame long feature track shows in the extended Kalman filter, where the estimated feature location must be summarized into a multivariate Gaussian distribution. The black dots depict the true distribution calculated by a Monte Carlo scheme, the red patch show the 95\% confidence ellipses for the MSCKF visual update, and the blue patch the 95\% confidence ellipses for the PIVO visual update. The MSCKF approximation is coarse, but the approximate density covers the true one with high certainty. Taking all cross-correlation into account and successfully accounting for the sensitivity of the estimated feature location, makes the PIVO update model more accurate. The directions of the correlations (tilt of the distributions) are interpreted right. The nature of the local linearization can still keep the mean off.

When proposing a visual update, for robustness against outlier tracks, we use the standard chi-squared innovation test approach (see, \eg, \cite{Bar-Shalom+Li+Kirubarajan:2001}), which takes into account both the predicted visual track and the estimate uncertainty.

\begin{table*}
  \caption{Comparison results on the EuRoC MAV dataset in absolute trajectory error RMSE (median) in meters. The other tabulated results for other methods are collected from Krombach~\etal\ \cite{Krombach+Droeschel+Behnke:2016}${^\dagger}$ and Mur-Artal and Tard{\'{o}}s \cite{Mur-Artal+Tardos:2016}${^\ddagger}$, \cite{Mur-Artal+Tardos:2017b}*.}
  \label{tbl:euroc}
  \centering

\resizebox{\textwidth}{!}{%
  \begin{tabular}{ l c c c c c c c c } 
  \toprule
  Dataset & PIVO (ours) & HYBRID$^\dagger$ & LIBVISO2$^\dagger$ & LSD-SLAM$^\dagger$ & ORB-SLAM$^\dagger$ & S-PTAM$^\dagger$ & ORB-SLAM2$^\ddagger$ & VI-SLAM* \\ 
  \midrule
  V1 01 (easy) & 0.82 (0.31)  & 0.25 (0.18) & 0.31 (0.31) & 0.19 (0.10) & 0.79 (0.62) & 0.28 (0.19) & 0.035 & 0.027\\ 
  V1 02 (medium)& --- & 0.24 (0.16) & 0.29 (0.27) & 0.98 (0.92) & 0.98 (0.87) & 0.50 (0.35) & 0.020 & 0.028\\ 
  V1 03 (difficult) & 0.72 (0.48)  & 0.81 (0.76) & 0.87 (0.64) & --- & 2.12 (1.38) & 1.36 (1.09) & 0.048 & --- \\ 
  V2 01 (easy) & 0.11 (0.07) & 0.22 (0.13) & 0.40 (0.31) & 0.45 (0.41) & 0.50 (0.42) & 2.38 (1.78) & 0.037 & 0.032\\ 
  V2 02 (medium) & 0.24 (0.15) & 0.31 (0.25) & 1.29 (1.08) & 0.51 (0.48) & 1.76 (1.39) & 4.58 (4.18) & 0.035 & 0.041\\ 
  V2 03 (difficult) & 0.51 (0.23)  & 1.13 (0.97) & 1.99 (1.66) & --- & --- & --- & --- & 0.074\\ 
  \midrule
  Mean & 0.48 (0.25) & 0.49 (0.41) & 0.85 (0.71) & 0.53 (0.48) & 1.23 (0.94) & 1.82 (1.52) & 0.035 & 0.040\\
  \bottomrule
  \end{tabular}%
}

\end{table*}

\section{Results}
\label{sec:results}
In the following we present a number of experiments which aim to demonstrate the proposed method to be comparable with the current state-of-the-art, to provide added value in comparison to the Google Tango device, and to provide useful results in challenging real-world environments with smartphone hardware.

\subsection{EuRoC MAV data}
We compare the proposed PIVO method to other recent methods by using the publicly available EuRoC MAV dataset \cite{Burri+Nikolic+Gohl+Schneider+Rehder+Omari+Achtelik+Siegwart:2016}.
This dataset contains sequences of data recorded from an IMU and two synchronized video cameras on board a micro-aerial vehicle. 
We consider a subset of the data which contains ground-truth trajectories from a VICON system, and we only use camera~`0'.

\begin{figure}
  \centering\hspace{\fill}
  \begin{subfigure}[t]{.5\textwidth}
    \centering
    \includegraphics[width=.85\textwidth]{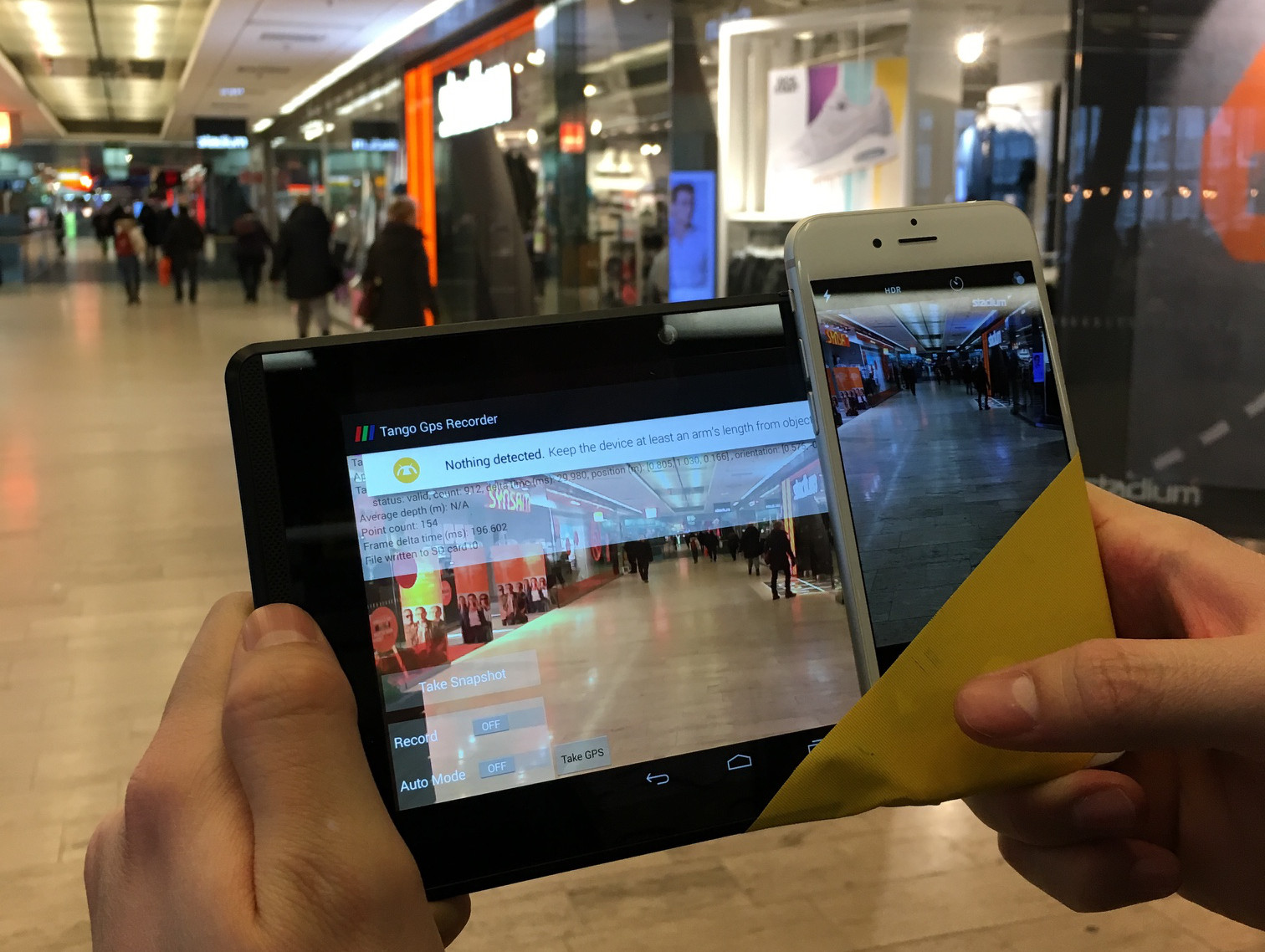}
    \caption{Experiment setup\\}%
    \label{fig:tango-iphone}%
  \end{subfigure}
  \\[1em]

  \setlength{\figurewidth}{.5\textwidth}
  \begin{subfigure}[t]{.5\textwidth}
    \footnotesize\centering

    \pgfplotsset{
      trim axis left,
      yticklabel style={rotate=90},
    }
    \footnotesize\centering
    \input{./fig/comparison-horizontal.tex}

    \caption{Effect of a complete occlusion \\ \ref{addplot:comparison-horizontal4} Start \ref{addplot:comparison-horizontal5} End \hspace{1em}\ref{addplot:comparison-horizontal3} Occluded}%
    \label{fig:comparison}%
  \end{subfigure}

  \caption{Comparison study between PIVO (iPhone~6) and a Tango device. The Tango is sensitive to occlusions and failed to return to the starting point at the end.}
  \vspace{-1em}
\end{figure}

The EuRoC MAV sequences contain typically several passes through the same scene and thus are best suited for SLAM approaches where a map is built on the fly, or methods employing automatic loop-closures. The inertial information contains very pronounced high-frequency noise from the rotors and relatively soft movement slowing down sensor bias estimation. We present our results on the dataset to show that our system performs on par with recently published methods despite this use not being its core purpose.

Our results are shown in Table~\ref{tbl:euroc}. The values for other methods are obtained from \cite{Krombach+Droeschel+Behnke:2016, Mur-Artal+Tardos:2016, Mur-Artal+Tardos:2017b}. All other methods in Table~\ref{tbl:euroc}, except ours and VI-SLAM \cite{Mur-Artal+Tardos:2017b}, are visual SLAM methods for monocular or stereo cameras. The results for LIBVISO2, LSD-SLAM, S-PTAM, and ORB-SLAM have been computed by Krombach~\etal\ using open-source software from original authors with default parameters \cite{Krombach+Droeschel+Behnke:2016}. Our results are comparable with \cite{Krombach+Droeschel+Behnke:2016} but worse than in \cite{Mur-Artal+Tardos:2016,Mur-Artal+Tardos:2017b}. While both \cite{Mur-Artal+Tardos:2017b} and our method use inertial sensors, ours is not a SLAM approach, that is we do not utilize loop-closures or re-localization like \cite{Mur-Artal+Tardos:2016,Mur-Artal+Tardos:2017b}.

It is important to note that map reuse is highly beneficial in the case of EuRoC MAV datasets since the camera moves in a small environment only some meters in size. However, the situation would be different in use cases where the device moves long distances without re-visiting previous areas. In fact, instead of SLAM methods, our results should be compared to other visual-inertial odometry approaches, such as \cite{Usenko+Engel+Stuckler+Cremers:2016}. Unfortunately, \cite{Usenko+Engel+Stuckler+Cremers:2016} does not report the global RMS error on EuRoC data and detailed comparison is therefore not possible.

\subsection{Demonstration of robustness to occlusion}
In the following experiments, we use an off-the-shelf Apple iPhone~6. The data acquisition was implemented in Objective-C, while the reconstruction was done off-line in Mathworks Matlab with performance critical components written in C++. The IMU data is sampled at 100~Hz. The experiments use the rear-facing camera at 10~fps, with a resolution of $480 \times 640$ (portrait orientation), grayscale images, shutter speed $1/60$, an ISO value of 200, and locked focus. The IMU and camera data were acquired time-synced on the device. The camera distortion parameters were estimated off-line prior to the data acquisition.

There are not many publicly available visual-inertial odometry methods, and none of them is shown to be robust to occlusion and to work with standard MEMS-based inertial sensors and rolling shutter cameras of current smartphones (namely those of iPhone~6) in challenging conditions (\eg\ with rapid motions and moving obstacles). Therefore, the proprietary visual-inertial odometry technology of the Google Tango device was the only baseline which we could use for comparisons. We attached the iPhone and the Tango device together, and captured similar motion trajectories with both devices. We observed that both approaches, PIVO and Tango, produced comparable and accurate trajectories in normal capture conditions without occlusions, but only PIVO was robust to occlusion.

During data acquisition the iPhone was firmly attached to the Tango in order to keep the tracks aligned (see Fig.~\ref{fig:tango-iphone}). We ran the built-in visual-inertial odometry method on the Tango. Figure~\ref{fig:comparison} shows an example of estimated tracks both for PIVO and the Tango device using data acquired in a shopping mall. After walking for some 110~m, the camera was completely occluded for the dashed part of the path. While the Tango track exhibits a discontinuity at the occlusion, PIVO is able to track movement even though being completely occluded. At the end of the path PIVO is horizontally off from the starting point 29~cm (0.23\% rel.\ error to traveled distance).

\begin{figure}

  \footnotesize\centering

  \setlength{\figurewidth}{.5\textwidth}
  \input{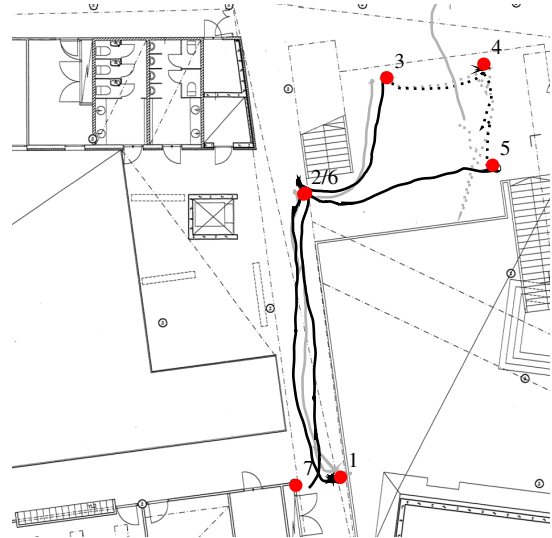}
  \caption{Occlusion experiment: A motion sequence through seven known points (\ref{addplot:pivo-and-tango0}) was recorded with an iPhone~6 and Tango device so that the cameras were occluded between points 3 and 5 (dashed lines). The trajectories estimated by the Tango device (\ref{addplot:pivo-and-tango1}) and our PIVO method (\ref{addplot:pivo-and-tango2}) from iPhone data are shown. Tracking on the Tango device fails due to occlusion but the PIVO path is still accurate (RMS error 0.74 meters, also confirmed by visual inspection).}
  \label{fig:supplementary}
  \vspace*{-1em}
\end{figure}

In the second occlusion experiment, which is also illustrated in Figure~\ref{fig:supplementary} above. In this experiment, the Tango device was again rigidly attached to the iPhone~6, and a walk was performed between seven known points in a building. The points were close to supporting pillars of the building and their location was precisely determined from the architectural floor plan. The camera lenses of both Tango and iPhone were occluded by hand simultaneously between points 3 and 5. The original iPhone video and the PIVO result are shown in the supplementary video. At each known point the devices were firmly placed to the ground truth location so that the corresponding time was easy to recognise from the stationary part in sensor recordings. The estimated PIVO trajectory was aligned onto the floorplan image by fitting a least-squares 2D rotation and translation that maps the seven points of the trajectory to the floor plan. The residual RMS distance to the known points was 0.74~meters after fitting and the result is illustrated in Figure~\ref{fig:supplementary}. Due to the occlusion the tracking of the Tango device failed completely between points 4 and 5 as illustrated in Figure~\ref{fig:supplementary}. Thus, the Tango trajectory was aligned onto the floorplan image using just the first four points before divergence.

Without occlusions, Tango and PIVO performances are close to each other. In general, the fisheye lens in the Tango gives it an upper hand, but with enough movement for exciting the sensors with the narrow field of view in the iPhone, PIVO keeps up. We performed an experiment where a set of seven ground-truth points were visited along a 68~meter path. The Tango trajectory gave an RMSE of 0.26~m, while PIVO gave 0.59~m. This suggests that comparable results to the Tango can be gained without its specialized hardware.

\subsection{Tracking a smartphone through city center}
Figure~\ref{fig:example} shows a large-scale odometry example on an iPhone, where the path is started by descending from the fourth floor of an office building, then walking through city streets to a shopping mall, descending to a metro station, and finally re-surfacing. Ill-lit indoor spaces, direct sunlight, occluding crowds, passing cars, and feature-poor environments make the visual environment challenging. To better balance the camera settings between indoor and outdoor, we used a shutter speed of $1/1000$, ISO value 400, and $n_\mathrm{a}=40$. The reconstructed path, about 600~m in length, is overlaid on a city map. A video demonstrating the reconstruction is included in the supplementary material\footnote{Also available at \url{http://arno.solin.fi}.}.

The supplementary video shows the original iPhone video and PIVO result for the experiment illustrated in Figure~\ref{fig:example}. The captured image data contains many challenges that usually hamper tracking, such as low-light illumination conditions, over-exposure due to direct sunlight, and moving people in front of the camera, which cause that several times during the course of the trajectory there are effectively no visual feature tracks or most of them are not compatible with the device motion.

For example, these kind of situations happen in the beginning (at 01:14  from the start of the video) and in the end of the video (at 05:35 from the start). Further, the frame rate is relatively low (10 frames per second) and there are rapid motions causing motion blur and possibly also rolling shutter effects. Despite the aforementioned challenges PIVO performs accurately and robustly during the whole path of several hundreds of meters. The visual inspection, performed after aligning the trajectory onto the map image by rigid transformation (Fig.~\ref{fig:example}), shows that the estimated trajectory matches the street map very accurately (including scale).

It should be noted that most of the solely camera-based odometry methods will not work in cases like the one illustrated in the attached video. In fact, we tried the publicly available open-source implementation of ORB-SLAM2 for several of our iPhone videos and could not get it working in such challenging cases as those shown in our attached videos. However, ORB-SLAM2 worked fine when the camera motion was less rapid and when the camera was observing mostly the same small scene during the entire trajectory (\ie\ not moving constantly and rapidly forward to new previously unseen areas).

\section{Discussion and conclusion}
\label{sec:discussion}
In this paper we have derived a method for fusing inertial odometry with visual data. These two modalities are complementary in the sense that the inertial data provides fast-sampled high-accuracy estimates in short time-windows, while the visual data provide relative info of movement between consecutive frames. Therefore the visual data is ideal for correcting long-term drifting and bias effects introduced by low-cost IMU sensors. On the other hand, the IMU data is completely occlusion and use-case invariant, while the visual data is sensitive to occlusions and feature-poor environments.

Our PIVO method is able produce impressive results in challenging visual environments with few or no features (occlusion), rapid motion, changing lighting conditions, rolling shutter, and small field-of-view. We are not aware of previous work able to do this on a standard smartphone.

There are three key factors in the presented PIVO method. First, we formulate the IMU propagation model directly in discrete-time, which gains us some efficiency. Second, the trailing pose augmentation is accomplished through a series of linear Kalman update tricks which preserve the cross-covariance between past and future poses. Third, the visual update resembles that of MSCKF \cite{Mourikis+Roumeliotis:2007}, but as they mention at the end of their appendix: ``\emph{\dots the camera pose estimates are treated as known constants, and their covariance matrix is ignored. As a result, the minimization can be carried out very efficiently, at the expense of the optimality of the feature position estimates}''. We on the other hand take all the cross-terms into account, and our scheme is optimal in the first-order EKF linearization sense. Therefore we call our method `probabilistic'.

The sequential formulation of PIVO makes the computational complexity scale linearly with respect to time. Computationally the IMU propagation and pose augmentation updates are extremely fast on an iPhone and negligible with respect to the total computation time. Most of the computation time is spent in feature tracking (OpenCV) and the visual update. The slowest separate part is calculating the visual update (including derivatives of Gauss-Newton). Benchmarks (on a MacBook laptop) of our implementation show that calculating the visual update takes on average 0.079~s (for a visual track of 40 frames) or 0.0052~s (for a track of 20 frames). The algorithm is capable of running on, for example, any recent iPhone.

A computational detail of the PIVO method is that in terms of the last augmented pose in the state, our model gives the same result as a fixed-lag extended RTS smoother would give: the pose estimate conditioned on all the data seen up to the most recent augmented pose (see, \eg, \cite{Kaess+Johannsson+Roberts+Ila+Leonard+Dellaert:2012} for smoothing methods for visual odometry).

In the scope of this paper, we only consider sequential odometry, and the methodology could be extended with loop-closures which would bring it closer to recently published SLAM methods, and help with slow crawling. On the other hand, SLAM methods are only beneficial if the same location is re-visited several times, and in cases such as Figure~\ref{fig:example} SLAM would not bring any added value. The current sequential formulation of PIVO also means that computational complexity does not grow over time. Furthermore, visual-heavy methods perform poorly on narrow field-of-view cameras such as the one in iPhones. Also the PIVO method could be extended to perform even better on low-cost camera hardware---\eg\ by taking rolling-shutter effects into account. Other future directions include delayed pose disposal for loop-closures and fusing additional data (barometric height, GPS locations) with the model.

\section*{Acknowledgements}
Authors acknowledge funding from the Academy of Finland (grant numbers 277685, 295081, 308640, 310325).

\bibliographystyle{ieee}

{\small
\bibliography{bibliography}

\begin{thebibliography}{10}\itemsep=-1pt

\bibitem{Bar-Shalom+Li+Kirubarajan:2001}
Y.~Bar-Shalom, X.-R. Li, and T.~Kirubarajan.
\newblock {\em Estimation with Applications to Tracking and Navigation}.
\newblock Wiley-Interscience, New York, 2001.

\bibitem{Blosch+Omari+Hutter+Siegwart:2015}
M.~Bl{\"{o}}sch, S.~Omari, M.~Hutter, and R.~Siegwart.
\newblock Robust visual inertial odometry using a direct {EKF}-based approach.
\newblock In {\em Proceedings of the International Conference on Intelligent
  Robots and Systems (IROS)}, pages 298--304, Hamburg, Germany, 2015.

\bibitem{Bouguet:2001}
J.-Y. Bouguet.
\newblock Pyramidal implementation of the affine {L}ucas {K}anade feature
  tracker: {D}escription of the algorithm.
\newblock Technical report, Intel Corporation, 2001.

\bibitem{Burri+Nikolic+Gohl+Schneider+Rehder+Omari+Achtelik+Siegwart:2016}
M.~Burri, J.~Nikolic, P.~Gohl, T.~Schneider, J.~Rehder, S.~Omari, M.~W.
  Achtelik, and R.~Siegwart.
\newblock The {EuRoC} micro aerial vehicle datasets.
\newblock {\em International Journal of Robotics Research}, 35:1157--1163,
  2016.

\bibitem{Engel+Schops+Cremers:2014}
J.~Engel, T.~Sch{\"{o}}ps, and D.~Cremers.
\newblock {LSD-SLAM:} {L}arge-scale direct monocular {SLAM}.
\newblock In {\em Proceedings of European Conference on Computer Vision
  (ECCV)}, pages 834--849, Zurich, Switzerland, 2014.

\bibitem{Engel+Stuckler+Cremers:2015}
J.~Engel, J.~St{\"{u}}ckler, and D.~Cremers.
\newblock Large-scale direct {SLAM} with stereo cameras.
\newblock In {\em Proceedings of the International Conference on Intelligent
  Robots and Systems (IROS)}, pages 1935--1942, Hamburg, Germany, 2015.

\bibitem{Forster+Carlone+Dellaert+Scaramuzza:2017}
C.~Forster, L.~Carlone, F.~Dellaert, and D.~Scaramuzza.
\newblock On-manifold preintegration for real-time visual--inertial odometry.
\newblock {\em Transactions on Robotics}, 33(1):1--21, 2017.

\bibitem{Forster+Pizzoli+Scaramuzza:2014}
C.~Forster, M.~Pizzoli, and D.~Scaramuzza.
\newblock {SVO}: {F}ast semi-direct monocular visual odometry.
\newblock In {\em Proceedings of the International Conference on Robotics and
  Automation (ICRA)}, pages 15--22, Hong Kong, China, 2014.

\bibitem{Forster+Zhang+Gassner+Werlberger+Scaramuzza:2017}
C.~Forster, Z.~Zhang, M.~Gassner, M.~Werlberger, and D.~Scaramuzza.
\newblock {SVO}: {S}emidirect visual odometry for monocular and multicamera
  systems.
\newblock {\em Transactions on Robotics}, 33(2):249--265, 2017.

\bibitem{Heikkila+Silven:1997}
J.~Heikkila and O.~Silven.
\newblock A four-step camera calibration procedure with implicit image
  correction.
\newblock In {\em Proceedings of the Computer Society Conference on Computer
  Vision and Pattern Recognition (CVPR)}, pages 1106--1112, San Juan, Puerto
  Rico, 1997.

\bibitem{Heng+Choi:2016}
L.~Heng and B.~Choi.
\newblock Semi-direct visual odometry for a fisheye-stereo camera.
\newblock In {\em Proceedings of International Conference on Intelligent Robots
  and Systems (IROS)}, pages 4077--4084, Daejeon, Korea, 2016.

\bibitem{Herrera+Kim+Kannala+Pulli+Heikkila:2014}
D.~Herrera, K.~Kim, J.~Kannala, K.~Pulli, and J.~Heikkil{\"{a}}.
\newblock {DT-SLAM:} {D}eferred triangulation for robust {SLAM}.
\newblock In {\em International Conference on 3D Vision (3DV)}, pages 609--616,
  Tokyo, Japan, 2014.

\bibitem{Hesch+Kottas+Bowman+Roumeliotis:2014}
J.~A. Hesch, D.~G. Kottas, S.~L. Bowman, and S.~I. Roumeliotis.
\newblock Consistency analysis and improvement of vision-aided inertial
  navigation.
\newblock {\em Transactions on Robotics}, 30(1):158--176, 2014.

\bibitem{Izadi+Newcombe+Kim+Hilliges+Molyneaux+Hodges+Kohli+Shotton+Davison+Fitzgibbon:2011}
S.~Izadi, R.~A. Newcombe, D.~Kim, O.~Hilliges, D.~Molyneaux, S.~Hodges,
  P.~Kohli, J.~Shotton, A.~J. Davison, and A.~W. Fitzgibbon.
\newblock {KinectFusion}: {R}eal-time dynamic {3D} surface reconstruction and
  interaction.
\newblock In {\em Proceedings of the International Conference on Computer
  Graphics and Interactive Techniques (SIGGRAPH)}, page~23, Vancouver, BC,
  Canada, 2011.

\bibitem{Jazwinski:1970}
A.~H. Jazwinski.
\newblock {\em Stochastic Processes and Filtering Theory}.
\newblock Academic Press, New York, 1970.

\bibitem{Kaess+Johannsson+Roberts+Ila+Leonard+Dellaert:2012}
M.~Kaess, H.~Johannsson, R.~Roberts, V.~Ila, J.~J. Leonard, and F.~Dellaert.
\newblock {iSAM2}: {I}ncremental smoothing and mapping using the {B}ayes tree.
\newblock {\em International Journal of Robotics Research}, 31(2):216--235,
  2012.

\bibitem{Kerl+Stuckler+Cremers:2015}
C.~Kerl, J.~St{\"{u}}ckler, and D.~Cremers.
\newblock Dense continuous-time tracking and mapping with rolling shutter
  {RGB-D} cameras.
\newblock In {\em Proceedings of the International Conference on Computer
  Vision (ICCV)}, pages 2264--2272, Santiago, Chile, 2015.

\bibitem{Klein+Murray:2009}
G.~Klein and D.~W. Murray.
\newblock Parallel tracking and mapping on a camera phone.
\newblock In {\em International Symposium on Mixed and Augmented Reality
  (ISMAR)}, Science {\&} Technology Proceedings, pages 83--86, Orlando, FL,
  USA, 2009.

\bibitem{Krombach+Droeschel+Behnke:2016}
N.~Krombach, D.~Droeschel, and S.~Behnke.
\newblock Combining feature-based and direct methods for semi-dense real-time
  stereo visual odometry.
\newblock In {\em International Conference on Intelligent Autonomous Systems
  (IAS)}, pages 855--868, Shanghai, China, 2016.

\bibitem{Leutenegger+Lynen+Bosse+Siegwart+Furgale:2015}
S.~Leutenegger, S.~Lynen, M.~Bosse, R.~Siegwart, and P.~Furgale.
\newblock Keyframe-based visual--inertial odometry using nonlinear
  optimization.
\newblock {\em International Journal of Robotics Research}, 34(3):314--334,
  2015.

\bibitem{Li+Kim+Mourikis:2013}
M.~Li, B.~Kim, and A.~I. Mourikis.
\newblock Real-time motion tracking on a cellphone using inertial sensing and a
  rolling shutter camera.
\newblock In {\em Proceedings of the International Conference on Robotics and
  Automation (ICRA)}, pages 4712--4719, Karlsruhe, Germany, 2013.

\bibitem{Lucas+Kanade:1981}
B.~D. Lucas and T.~Kanade.
\newblock An iterative image registration technique with an application to
  stereo vision.
\newblock In {\em Proceedings of the International Conference on Artificial
  Intelligence (IJCAI)}, pages 674--679. Vancouver, BC, Canada, 1981.

\bibitem{Montiel+Civera+Davison:2006}
J.~M.~M. Montiel, J.~Civera, and A.~J. Davison.
\newblock Unified inverse depth parametrization for monocular {SLAM}.
\newblock In {\em Proceedings of Robotics: Science and Systems}, Philadelphia,
  USA, 2006.

\bibitem{Mourikis+Roumeliotis:2007}
A.~I. Mourikis and S.~I. Roumeliotis.
\newblock A multi-state constraint {K}alman filter for vision-aided inertial
  navigation.
\newblock In {\em Proceedings of the International Conference on Robotics and
  Automation (ICRA)}, pages 3565--3572, Rome, Italy, 2007.

\bibitem{Mur-Artal+Tardos:2016}
R.~Mur{-}Artal and J.~D. Tard{\'{o}}s.
\newblock {ORB-SLAM2:} {A}n open-source {SLAM} system for monocular, stereo and
  {RGB-D} cameras.
\newblock {\em arXiv preprint arXiv:1610.06475}, 2016.

\bibitem{Mur-Artal+Tardos:2017b}
R.~Mur-Artal and J.~D. Tard{\'o}s.
\newblock Visual-inertial monocular {SLAM} with map reuse.
\newblock {\em Robotics and Automation Letters}, 2(2):796--803, 2017.

\bibitem{Newcombe+Fox+Seitz:2015}
R.~A. Newcombe, D.~Fox, and S.~M. Seitz.
\newblock {DynamicFusion}: {R}econstruction and tracking of non-rigid scenes in
  real-time.
\newblock In {\em {IEEE} Conference on Computer Vision and Pattern Recognition
  (CVPR)}, pages 343--352, Boston, MA, USA, 2015.

\bibitem{Sachs:2010}
D.~Sachs.
\newblock Sensor fusion on {A}ndroid devices: {A} revolution in motion
  processing.
\newblock \url{https://www.youtube.com/watch?v=C7JQ7Rpwn2k}, 2010.
\newblock [Online; accessed 17-March-2017].

\bibitem{Sarkka:2013}
S.~S{\"a}rkk{\"a}.
\newblock {\em Bayesian Filtering and Smoothing}.
\newblock Institute of Mathematical Statistics Textbooks. Cambridge University
  Press, 2013.

\bibitem{Sarkka+Solin:2012}
S.~S\"arkk\"a and A.~Solin.
\newblock On continuous-discrete cubature {K}alman filtering.
\newblock In {\em Proceedings of {SYSID} 2012, 16th {IFAC} Symposium on System
  Identification}, pages 1210--1215, Brussels, Belgium, 2012.

\bibitem{Shi+Tomasi:1994}
J.~Shi and C.~Tomasi.
\newblock Good features to track.
\newblock In {\em Proceedings of the IEEE Computer Society Conference on
  Computer Vision and Pattern Recognition (CVPR)}, pages 593--600, 1994.

\bibitem{Solin+Cortes+Rahtu+Kannala}
A.~Solin, S.~Cortes, E.~Rahtu, and J.~Kannala.
\newblock Inertial odometry on handheld smartphones.
\newblock {\em arXiv preprint arXiv:1703.00154}, 2017.

\bibitem{Strasdat+Montiel+Davison:2010}
H.~Strasdat, J.~M.~M. Montiel, and A.~J. Davison.
\newblock Scale drift-aware large scale monocular {SLAM}.
\newblock In {\em Proceedings of Robotics: Science and Systems}, Zaragoza,
  Spain, 2010.

\bibitem{Tanskanen+Kolev+Meier+Camposeco+Saurer+Pollefeys:2013}
P.~Tanskanen, K.~Kolev, L.~Meier, F.~Camposeco, O.~Saurer, and M.~Pollefeys.
\newblock Live metric {3D} reconstruction on mobile phones.
\newblock In {\em Proceedings of the International Conference on Computer
  Vision (ICCV)}, pages 65--72, Sydney, Australia, 2013.

\bibitem{Tanskanen+Naegeli+Pollefeys+Hilliges:2015}
P.~Tanskanen, T.~Naegeli, M.~Pollefeys, and O.~Hilliges.
\newblock Semi-direct {EKF}-based monocular visual-inertial odometry.
\newblock In {\em Proceedings of the International Conference on Intelligent
  Robots and Systems (IROS)}, pages 6073--6078, Hamburg, Germany, 2015.

\bibitem{Titterton+Weston:2004}
D.~H. Titterton and J.~L. Weston.
\newblock {\em Strapdown Inertial Navigation Technology}.
\newblock The Institution of Electrical Engineers, 2004.

\bibitem{Usenko+Engel+Stuckler+Cremers:2016}
V.~Usenko, J.~Engel, J.~St{\"u}ckler, and D.~Cremers.
\newblock Direct visual-inertial odometry with stereo cameras.
\newblock In {\em Proceedings of the International Conference on Robotics and
  Automation (ICRA)}, pages 1885--1892, Stockholm, Sweden, 2016.

\end{thebibliography}
}

\end{document}